%% file: main.tex
\documentclass{article}
\usepackage{spconf,amsmath,graphicx}
\usepackage{cite}
\usepackage{color, colortbl}
\definecolor{LightCyan}{rgb}{0.88,1,1}

\usepackage[ruled]{algorithm2e} 

\title{MABNet: Master Assistant Buddy Network with Hybrid Learning for Image Retrieval}
\name{Rohit Agarwal$^{\dagger\ast}$ , Gyanendra Das$^{\ddagger\ast}$, Saksham Aggarwal$^\ddagger$\sthanks{Equal contribution}, Alexander Horsch$^\dagger$, Dilip K. Prasad$^\dagger$}
\address{$^\dagger$Bio-AI Lab, UiT The Arctic University of Norway, Tromsø, Norway \\
$^\ddagger$Indian Institute of Technology (Indian School of Mines), Dhanbad, India}
\begin{document}
%
\maketitle
\begin{abstract}
Image retrieval has garnered growing interest in recent times. The current approaches are either supervised or self-supervised. These methods do not exploit the benefits of hybrid learning using both supervision and self-supervision. We present a novel Master Assistant Buddy Network (MABNet) for image retrieval which incorporates both learning mechanisms. MABNet consists of master and assistant blocks, both learning independently through supervision and collectively via self-supervision. The master guides the assistant by providing its knowledge base as a reference for self-supervision and the assistant reports its knowledge back to the master by weight transfer. We perform extensive experiments on public datasets with and without post-processing.
\end{abstract}
\begin{keywords}
Image Retrieval, Supervision, Self-Supervision, MABNet, ViT
\end{keywords}

\input{Figures/Architecture}

\section{Introduction}
\label{sec:intro}
Image retrieval refers to the task of returning relevant instances from a database given an unlabeled query image. This task can be targeted in both supervised and self-supervised manners. Supervised methods form better decision boundaries owing to the ground truth and are commonly based on Convolutional Neural Networks (CNN) like ResNets \cite{DBLP:journals/corr/ToliasSJ15, radenovic2016cnn, gordo2016deep, noh2017large, hoang2017selective, cao2020unifying, yang2021dolg, song2022dalg, hu2022expressive, Lee_2022_CVPR, revaud2019learning, ng2020solar} since they can extract image-level descriptors.

Recently, self-supervised learning has emerged to address the image retrieval task \cite{caron2021emerging, elnouby2021training}. DINO\cite{caron2021emerging} exploits the attention mechanism of Vision Transformer (ViT) for self-supervision such that its knowledge model contains explicit information about the semantic segmentation of an image.

Supervision and self-supervision have their own advantages and the current literature has not explored the use of both in image retrieval task. Self-supervision will complement the supervised approach with explicit semantic segmentation improving the decision boundaries.  
We propose Master Assistant Buddy Network (MABNet), a novel model that employs both supervised and self-supervised learning mechanisms, thereby incorporating the advantages of both learning paradigms. MABNet is a two-block buddy network (see Fig. \ref{fig:architecture}), where one is master and the other is assistant. Both learn individually via supervised learning. In addition, they compare their latent features for self-supervision using a distance metric where the assistant uses the master's latent features as a reference for self-supervision. The weights of the master block are later updated by the assistant's knowledge model via weight transfer. In this sense, the assistant network shares the learning load, performs self-supervision, and assists the master with a comprehensive knowledge model. Further, the master and the assistant divide the focus of learning where the master uses both global and local image crops, whereas the assistant specializes in global crops.

Convolution is a local operator whereas transformer is a global operator that considers all the pixels and applies attention to find the important features. It is known that effective receptive fields of lower layers for ViTs are much larger than ResNets, which helps them to incorporate more global information than CNNs\cite{caron2021emerging,elnouby2021training}. Hence we employ ViT for both the master and the assistant blocks.

Several methods apply various post-processing techniques like Average Query Expansion (AQE) \cite{chum2011total}, heat diffusion\cite{pang2018deep}, offline diffusion \cite{yang2019efficient} and CVNet-Rerank\cite{Lee_2022_CVPR} to obtain the final solution by generating a ranked list of similar images. Thus we classify the image retrieval models into two categories: model without post-processing techniques and model with post-processing techniques. Some models \cite{cao2020unifying, yang2021dolg, song2022dalg, hu2022expressive} also include post-processing as part of their end-to-end pipeline, thus we classify them in the latter category.

\textbf{\textit{Contributions}} (1) We introduce the MABNet framework, which exploits both the supervised and the self-supervised learning paradigm. (2) We propose the use of the master and the assistant block with the master focusing on general knowledge learning and the assistant helping the master to refine its knowledge base through weight transfer. (3) We categorize all the image retrieval models either with or without post-processing approaches. We train MABNet on the GLDv2-train-clean (GLDv2-TC)\cite{weyand2020google} dataset and demonstrate its efficacy on popular benchmark datasets, namely Oxford (Ox5k and Ox105k)\cite{philbin2007object}, Paris (P6k and P106k)\cite{philbin2008lost}, Revisited Oxford (ROx(M) and ROx(H))\cite{radenovic2018revisiting} and Revisited Paris (RPar(M) and RPar(H))\cite{radenovic2018revisiting} in both the categories.

\section{Related Work}
CNN-based representation is an appealing solution for image retrieval. Early approaches used fully connected (FC) layer activations as global image descriptors. However, the FC layer does not learn geometric invariance and suppresses spatial information. Consequently, the last convolutional layer before the FC layer can be considered as containing the spatial context. MAC \cite{razavian2016visual} incorporated geometric invariance by choosing the last convolutional layer and adapting max-pooling, which boosted the model's performance on image retrieval tasks. 
Sum-Pooled Convolution (SPoC) \cite{babenko2015aggregating} used sum pooling from the feature map.
Ref. \cite{kalantidis2016cross} proposed non-parametric weighting schemes including cross-dimensional weighting and pooling steps. R-MAC \cite{DBLP:journals/corr/ToliasSJ15} used a hybrid approach by performing max-pooling over regions and sum-pooling of the regional descriptors. 
Generalized mean pooling (GEM) \cite{berman2019multigrain}, widely used for image retrieval tasks, generalizes max and average pooling with a pooling parameter. The combination of multiple global descriptors (CGD) \cite{jun2019combination} creates an ensemble effect by concatenating multiple global descriptors (including SPoC, MAC and GEM). Henceforth, CGD realizes a powerful hybrid approach for image retrieval.

Among the many CNN-based image retrieval techniques,  DELF \cite{noh2017large} and DIR \cite{gordo2016deep} have given state-of-the-art results. DELF is a local feature descriptor coupled with an attention mechanism
to identify semantically useful features for image retrieval. On the other hand, DIR produces a global and compact fixed-length representation for each image by aggregating many region-wise descriptors. It uses a three-stream Siamese network \cite{hoffer2015deep} that explicitly optimizes the weights of R-MAC representation for image retrieval by using triplet ranking loss. It also learns the pooling mechanism of the R-MAC descriptor. Selective deep conv features \cite{hoang2017selective} proposed the use of various masking strategies to select a subset of representative local convolutional features and boost its features' discriminative power by making use of the SOTA embedding methods. siaMAC \cite{radenovic2016cnn} is an unsupervised method of finetuning CNNs for image retrieval which exploits the 3D reconstructions to select the training data for CNN.

It is possible to embed the convolutional feature maps into a high dimensional space to obtain compact features, useful for image retrieval tasks. Widely used embedding methods include Bag of Words (BoW) \cite{1238663}, Vector of Locally Aggregated Descriptors (VLAD) \cite{5540039}, and Fischer vectors \cite{5540009}. VLAD captures information about the statistics of local descriptors aggregated over the image. BoW aggregation keeps the count of visual words, whereas VLAD stores the sum of residuals (difference between the descriptor and its corresponding cluster center) for each visual word. NetVLAd \cite{arandjelovic2016netvlad} is a generalized VLAD layer which contains trainable parameters. 

However, convolution is a local operator whereas the transformer is a global operator that considers all the pixels and applies attention to find the important features. In a study \cite{raghu2021vision}, it is shown that effective receptive fields of lower layers for ViTs are much larger than ResNets, which helps them to incorporate more global information than CNNs. Even though ViT lacks the inductive biases present in CNNs through translation equivariance and locality, it can give competitive or even better results on various tasks like classification, object detection, object tracking, and segmentation. Recently, some works \cite{caron2021emerging, elnouby2021training} used ViT for effective image retrieval. DINO's \cite{caron2021emerging}  knowledge model contains explicit information about the semantic segmentation of an image giving better performance. These reasons motivated us to use ViT as the backbone of our model. 

Some works also focus on designing loss functions specific to image retrieval tasks. \cite{ng2020solar} combines the second-order spatial attention and the second-order descriptor loss to improve image features for retrieval and matching. \cite{brown2020smooth} introduces an objective that optimizes Smooth-AP (smoothed approximation of average precision), instead of directly optimizing AP, which is non-differentiable.

\input{Algorithm/Algo1}
In order to increase the accuracy of image search systems, many post-processing techniques are employed as well. Query expansion (QE)~\cite{6248018,5995601, inproceedings} is a commonly used technique to achieve this goal, where relevant candidates produced during an initial ranking are aggregated into an expanded query, thus providing a robust representation of the query image, which is then used to search more images in the database. Aggregating the candidates reinforces the information shared between them and injects new information not available in the original query. The work in \cite{zhang2020understanding} presents another post-processing technique based on a graph neural network. Image re-ranking with heat diffusion~\cite{pang2018deep} re-ranks several top-ranked images for a given query image by considering the query as a heat source which helps to avoid over-representation of repetitive features. A downside to diffusion is that it performs slowly compared to the naive KNN search. The work in \cite{yang2019efficient} overcomes this problem by pre-computing diffusion results of each element in the database, making the online search a simple linear combination on top of the KNN search. It also proposes using late truncation, i.e., applying truncation after normalization to achieve better performance.

\input{Algorithm/Algo2}
\section{Method}

The proposed MABNet (see Fig.~\ref{fig:architecture}) consists of two blocks: the master block and the assistant block. Both the blocks have the backbone of ViT (can be any architecture like ResNet) denoted by $V$ but with different parameters $\theta_m$ and $\theta_a$ respectively. The role of the master is to perform image retrieval while focusing on local spatial context but retaining global context as well and the role of the assistant is to assist the master block by providing global spatial information through weight sharing, thereby enriching the representations learned by the master.

In MABNet, instead of passing the original input image $x$ to the network directly, we create multiple crops of the input image using the multi-crop method. We refer to the crops covering less than 50\% of the original image as local crops (denoted by $C_L$), and all other crops as global crops ($C_G$).  The master receives all the crops, whereas only the global crops are passed through the assistant. For an input image $x$, we get two embedding $E_m$ and $E_a$ from the GEM layer \cite{berman2019multigrain} corresponding to the master and assistant blocks, respectively. The $E_m$ and $E_a$ are given by the following equation where $x_m \in \{C_G \cup C_L\}$ and $x_a \in C_G$.
\begin{align}
  E_m(x) = V_{\theta_m}(x_m) 
  \quad ; \quad   
  E_a(x) = V_{\theta_a}(x_a) ,
\end{align}

The Kullback–Leibler (KL) divergence loss \cite{Joyce2011} is calculated between the embeddings $E_m$ and $E_a$. A fully-connected linear layer transforms the embeddings to the corresponding output labels. We employ ArcFace \cite{deng2019arcface} to get the supervised loss of each block. The gradient of each block with respect to the ArcFace loss flows independently and with respect to KL loss flows collectively. Further, we employ a weight-sharing technique from assistant to master, to update the master block's knowledge model. The final prediction is the output of the master since it captures more generalized information than the assistant. The architecture of MABNet is shown in Fig.~\ref{fig:architecture} and the pseudo algorithms for training and testing are given in Algorithms \ref{alg:one} and \ref{alg:two}. 

\subsection{Crops of Images}
We generate local and global crops from the original image and pass different combinations of image crops to each block. Using crops instead of the whole input image augments the data. The image of one object can be taken from different angles, resulting in small similarities between the images of the same object. Multi-crop strategy \cite{caron2020unsupervised} increases the chances of recognizing these images by comparing the embeddings of the parts of images. Besides, local crops impart contextual information to the master block, whereas the assistant block is rich with spatial information from global crops. 

\subsection{Loss functions used in our model training}
\label{sec:loss}
We discuss here the two losses used in MABNet, namely KL divergence loss \cite{Joyce2011} and ArcFace loss \cite{deng2019arcface}.

\textbf{KL divergence loss}
Since the two blocks are trained on different combinations of image crops, the embeddings of the blocks represent different features. The assistant block has spatial information, while the master block has more generalized embedding. We introduce a self-supervised KL divergence loss over the embeddings of both blocks. The master block embedding $E_m$ is used as the reference distribution since it has more generalized information. The KL divergence computes the relative entropy indicating how different the embedding $E_a$ is from $E_m$. The KL divergence loss for an input image $x$ is given by
\begin{align}
  L_{KL}(x) = \sum_{x_m \in \{C_G \cup C_L\}} \hspace{2mm}  \sum_{x_a \in C_G} E_m(x_m)log\frac{E_m(x_m)}{E_a(x_a)}
\label{eq:KL_loss}
\end{align}
where the notations are as defined before.
Through the KL divergence loss, we introduce a local-to-global correspondence, which helps the assistant learn more intricate information. This information is eventually passed back to the master block via the weight transfer method. Thus, the learning of both master and assistant goes hand-in-hand.

\textbf{ArcFace loss}
We use the supervised ArcFace loss in both blocks to have richer embeddings for the image retrieval task. The softmax loss does not explicitly enforce intra-class similarity and inter-class diversity. This deteriorates the performance for retrieval and recognition tasks, especially when there is high intra-class variations (as in \cite{moschoglou2017agedb}, \cite{zheng2017cross}, \cite{whitelam2017iarpa}). In contrast, ArcFace learns discriminative features by increasing the angular margin between classes. Further, it has a clear geometric interpretation due to the exact correspondence to the geodesic distance on the embedding hypersphere. The ArcFace transforms the logits, normalizes weights and embedding features to a unit magnitude, and rescales the embedding feature to a value $s$. Also, an additive angular margin penalty $m$ is added between the embedding feature $y_a^i$ and the ArcFace head weights $W_{y_i}$ for $y_i$-th class to enhance the intra-class compactness and inter-class discrepancy simultaneously. In our framework, these hyperparameters, $s$ and $m$ are learnable. The ArcFace loss of input image $x$ for the master block is given by
\begin{align}
  L_{AF}^M(x)=-\dfrac{1}{|C|}
  \sum_{i=1}^{|C|} log
        \dfrac{e^{s(cos(\theta_{y_i}+m))}}
              {e^{s(cos(\theta_{y_i}+m))}+              \sum_{j=1,j \neq y_i}^{n} e^{s cos(\theta_j)}}
\label{eq:ArcFace_loss}
\end{align}

where $C = \{C_G \cup C_L\}$ is the set of crops passed to the master block, $|\cdot|$ represents the cardinality, $n$ is the number of class labels, $W_j^Tx_i$ is the logit $cos(\theta_j)$ for each class, and $\theta_{y_i}$ is the angle between $y_a^i$ and the ground truth weight $W_{y_i}$. Similarly, the ArcFace loss of the assistant block $L_{AF}^A(x)$ follows the definition above, however, it uses the set of crops $C = C_G$.

\textbf{Total loss} The total loss of the model for a batch size of N is given by
\begin{align}
  L = \sum_{k=1}^{N} \{L_{KL}(x_k) + L_{AF}^M(x_k) + L_{AF}^A(x_k)\} 
\end{align}

\subsection{Weight Transfer} 
\label{sec:WeightsUpdate}

Both the blocks learn their weights by back-propagation via the two losses, self-supervised KL divergence loss, and supervised ArcFace loss. In addition to gradient descent, the weights of the master block are updated by the weighted average of master and assistant parameters as 
\begin{align}
  \theta_m = \lambda\theta_m + (1-\lambda)\theta_a
\label{eq:weight_transform}
\end{align}
where $\lambda$ is the weight transfer parameter. The value of $\lambda$ is determined by the grid approach and is in the range $0 \leq \lambda \leq 1$. 

\subsection{Testing Phase}
\label{sec:PostProcessingMethods}
For inference, we feed forward the input image and all its crops, i.e., $C_G$ and $C_L$ through the master block to get the embeddings coming from the GEM layer. Therefore, we have $C_G$+$C_L$+1 embeddings. We concatenate all of them. The embedding size for each image is 512. Then we train a simple KNN over these embeddings and get the nearest neighbors of the input image. Alternatively, we can use any post-processing techniques to get the final solution by generating a ranked list of similar images in the dataset. The most widely used post-processing techniques for image retrieval tasks are  Average Query Expansion (AQE) \cite{chum2011total}, heat diffusion\cite{pang2018deep}, offline diffusion \cite{yang2019efficient} and CVNet-Rerank\cite{Lee_2022_CVPR}.

\subsection{Model Working} 
Figure \ref{fig:imageretrievalexample} shows an example of the progressive learning of master and assistant blocks. The master embedding is more general and spread initially since it sees both the global and local crops. The assistant embedding is also spread but only to a part of the image depending on the global crops passed to it. But the assistant block compares its notes with the master block using self-supervision and hence learns the localized information. This can be seen by the heat maps in the later epochs as it becomes more focused on one particular area. The master block in turn gets this localized information from the assistant block in the form of the weight transfer and thus becomes more focused in the later epoch forming a better decision boundary. An example of images retrieved from the database is shown in the lower panel of Fig. \ref{fig:architecture} and Fig. \ref{fig:imageretrievalexample}(b). The embedding for a failure retrieval is also shown for insight.

\input{Figures/FigRetreivalExample}

\section{Datasets}
We trained the model (pre-trained on ImageNet) on Google Landmarks Dataset v2 (GLDv2) train-clean version and tested it on Oxford, Paris, Revisited Oxford, Revisited Paris and GLDv2 datasets.\\

\noindent\textbf{Oxford Buildings dataset}
We use both Oxford5k and Oxford105k datasets. Oxford5k is a collection of 5,062 images of the Oxford buildings \cite{Philbin07}. These images are collected from Flickr. It has 55 query images with 5 queries, each corresponding to a landmark and their respective ground truths. Oxford105k is the extension of the Oxford5k dataset, and it contains additionally 99782 negative images taken from Flickr \cite{Philbin08}.

\noindent\textbf{Paris dataset}
We report our result on both Paris6k and Paris106k of the Paris dataset \cite{Philbin08}. Paris6k consists of  6,392 high-resolution images of Paris city. Similar to Oxford5k, it is collected from Flickr by querying the associated text tags for famous Paris landmarks. This dataset also provides 55  query images and their corresponding ground truth images. Paris106k images is the extension of Paris6k datasets containing additional 100,000 Flickr images.

\noindent\textbf{Revisited Oxford and Paris datasets \cite{radenovic2018revisiting}} 
Oxford and Paris datasets were recently revised by addressing the annotation errors, the size of the dataset, and the level of challenge.  The revisited datasets, namely, ROxford and RParis, comprise 4,993 and 6,322 images respectively,  and a different query set for each, both with 70 images. The evaluation protocol is divided into Easy,
Medium and Hard difficulty levels. We report the results in challenging Medium and Hard categories. 

\noindent\textbf{Google Landmarks Dataset v2}
We use the train-clean version of Google Landmarks Dataset v2 (GLDv2) \cite{weyand2020google} which was released in a Kaggle challenge \textit{``Google Landmark Retrieval 2021"}. GLDv2 is a large-scale dataset, for instance-level recognition and retrieval task with over 5 million landmark images. GLDv2-train-clean~\cite{kim2021fairer} is a smaller, cleaner and more refined version of GLDv2, which is formed after certain pre-processing steps proposed in \cite{yokoo2020two} in order to make each class more visually coherent. The GLDv2-train-clean dataset contains 1,580,470 training images, 117,577 test images and 81,313 labels.

\section{Experiments}

\subsection{Comparison Methods}
We compare our method in two scenarios: its base form (without any post-processing method) and with post-processing of results. In its base form, we compare with state-of-the art methods like DIR \cite{gordo2016deep}, DELF \cite{noh2017large}, Deep Conv\cite{hoang2017selective}, siaMAC \cite{radenovic2016cnn}, R-MAC \cite{DBLP:journals/corr/ToliasSJ15}, Listwise Loss \cite{revaud2019learning}, IRT(R) \cite{elnouby2021training}, DINO \cite{caron2021emerging} and SOLAR \cite{ng2020solar}.
We then compare our model by applying the post-processing methods. Here, the comparison is done with DELG \cite{cao2020unifying}, DOLG \cite{yang2021dolg}, Swin-T-DALG \cite{song2022dalg}, Swin-S-DALG \cite{song2022dalg}, GeM \cite{hu2022expressive} and CVNet-Rerank \cite{Lee_2022_CVPR} which includes post-processing as a part of their pipeline.  

\subsection{Setup} 
We take 2 global and 8 local crops of random size. Random rescale is performed within the range [0.5, 1] and [0.05, 0.5] on global and local crops, respectively. The global and  local crops are resized to 224$\times$224 and 96$\times$96, respectively. We use ViT-small pre-trained on ImageNet as the backbone. Input patch size of 16$\times$16 is flattened and projected to $M$ lower dimensional linear vectors. The location prior is provided by adding trainable position encoding to the linear embedding before feeding them to the transformer encoder. The model is trained with 8$\times$V100s for 100 epochs with a batch size of 256. We use an AdamW optimizer with a learning rate of 0.0005 and a weight decay of $10^{-5}$. Furthermore, we use a linear warm-up for the first 10 epochs, after which the learning rate follows a cosine schedule. We use mixed-precision training to speed up computation. $\lambda$ is set as 0.5. We employ the standard mean average precision (mAP) metric to report our accuracy. The training time on GLDv2-TC \cite{weyand2020google} is 2.5 days. The number of flops is 9.2 GFlops. The inference time for all the concatenated embeddings of one image is 132 ms and for only the original image is 11 ms.

\input{Tables/WithoutPostProcessing}

\subsection{Without Post Processing}
\textbf{\textit{Oxford and Paris datasets}} The upper part of Table \ref{table:resultsw/opostprocessing} presents our results on the Oxford and Paris data. Our model surpasses the reported SOTA performances by a significant margin (6.1\% and 7.3\% better on the Ox105k and P106k datasets, respectively). The mAP score of our model is highest in Ox5k and P6k too. \textbf{\textit{Revisited datasets}}  MABNet outperforms other models by a margin of 15.1\% on ROx(M), 13.3\% on ROx(H), 3\% on RPar(M) and 8.4\% on RPar(H) (see Table \ref{table:resultsw/opostprocessing}).

\subsection{With Post-processing}
\textbf{\textit{Oxford and Paris datasets}} Cyan colored rows in Table \ref{table:resultswpostprocessing} show that the post-processing techniques improve the performance of MABNet. Offline diffusion provides the best improvement (3.9$-$7.6\%). MABNet is superior to other methods when applied to the same post-processing methods. \textbf{\textit{Revisited datasets}}  All the methods mentioned in the lower half of Table \ref{table:resultswpostprocessing} either uses the post-processing after the inference or they include post-processing technique as part of their model. Hence, we place them in the post-processing category. As seen from Table \ref{table:resultswpostprocessing}, MABNet with post-processing outperforms other methods in most cases.
\input{Tables/WithPostProcessing}

\subsection{Results on GLDv2 dataset}
We also applied our model to the test data of the GLDv2 dataset with and without post-processing. Since this is a relatively new dataset, we do not have many models to compare. Zhang et al. \cite{yuqi20212nd} reported an mAP score of 32.4\% whereas MABNet gave the mAP score of 37\% (see Table \ref{table:gldv2benchmarking}). Post-processing methods, especially offline diffusion further improved the model performance by 1.1\%.

\input{Tables/GLDv2Dataset}

\section{Ablation Studies}
We conducted a variety of ablation studies for MABNet on Ox5K dataset without post-processing (unless otherwise stated).

\subsection{Weight Transfer}  We considered training without weight transfer from the assistant to the master. The score dropped from 91.5\% to 81.9\%, which indicates its importance. We also investigated the effect of changing the direction of weight transfer, i.e., the assistant gets updated using weight transfer. This resulted in the drop of score to 80.5\%. We hypothesize that the weight transfer of master instead of assistant is more effective since the addition of global-only spatial information from assistant to the local-global contextual embedding of the master enriches it into a more generalized block.

\subsection{KL Divergence Loss} We consider the effect of dropping KL loss while training our model. The performance deteriorates from 91.3\% to 83.3\%, indicating that self-supervision is a critical determinant of MABNet. We also consider reversing the reference for KL loss, i.e., setting assistant embedding as a reference. The model performs poorer than MABNet (by 5.8\%), indicating the importance of the master as a reference. 

\subsection{Assistant block} We train our method with only the master block, removing the assistant block and all its dependent connections like weight transfer, KL and ArcFace loss. For a fair comparison, we consider all the crops and the original image. The score here drops to 80.2\% from 91.5\% for Ox5k, 77.6\% from 85.0\% for ROx(M) and 49.3\% from 61.2\% for ROx(H), indicating the importance of the assistant block.

\subsection{ViT backbone} We use ViT as the backbone due to its advantages (discussed in section \ref{sec:intro}). We replaced ViT with ResNet-50 and followed the same training procedure. Table \ref{table:resultsw/opostprocessing} shows that ViT gives significantly better performance than ResNet-50. Nonetheless, MABNet with ResNet-50 still performs better than other methods.

\subsection{Concatenated features} We use the original image, 2 global and 8 local crops for inference. The mAP score of MABNet on Ox5k is 91.5\%. Passing only the original image as input for inference gives 90.2\% which is still superior to the SOTA. The mAP scores were 85.7\% when only 2 global crops were passed and 85.4\% with only 8 local crops. Increasing the concatenated features with the original image, 4 global and 12 local crops gave mAP 91.7\% which is not a significant boost from 91.5\%. We conclude that the use of the original image, 2 global and 8 local crops gives a better balance between complexity and performance. Still, to decrease the inference time, one may use only the original images giving satisfactory results.

\input{Tables/Ensemble}

\subsection{Ensemble of two blocks}
MABNet uses only the master block to generate the final output. Here, we extend MABNet to MABNet(C), where we predict by concatenating the embeddings of the master and the assistant block. It is seen, MABNet(C) performs better in all the situations (see Table \ref{table:resultensemble}). MABNet is two times faster than MABNet(C) at inference time because it has half the number of parameters. The choice of MABNet and MABNet(C) is based on the trade-off between the inference time and the improved accuracy.

\input{Tables/LambdaAblation}
\subsection{Effect of $\lambda$ on MABNet}
The value of $\lambda$ determines the effect of weight transfer in MABNet. If $\lambda$ is 1, there is no effect of weight transfer on the model since no information is transferred from the assistant to the master. Whereas when the value is 0, the parameters of the master block are replaced by the assistant block, thus giving us exactly the same ViT block after each epoch. 

In our main experiment, we use the value of $\lambda$ = 0.5, thus making our weight transfer an average of master and assistant parameters. Here, we further optimize our method for different values of $\lambda$. We report our result on the Oxford5k dataset and Revisited Oxford and Paris datasets and show that the result of MABNet can be improved further on all the datasets upon extensive optimization. 
Here, we vary the values of $\lambda$ from 0.3 to 0.95 with a step size of 0.05. The result on the Oxford5k dataset for different $\lambda$ is shown in Table \ref{table:LambdaAblation}. 
It can be seen that the best result is observed at $\lambda$ = 0.6 (an increase of 1.7\%) as compared to the average value. The mAP score keeps increasing from $\lambda = 0.3$ (75.9 \%) to $\lambda = 0.6$ (93.2 \%), and after that it keeps on decreasing. When $\lambda$ is at its maximum value, i.e., $\lambda = 0.95$, we get an mAP score of 78.2\%. A similar trend in the scores can be seen on the revisited datasets as well where the maximum score is achieved when $\lambda$ lies between 0.55 and 0.65. Results are shown in Table \ref{table:LambdaAblation}. Thus, the performance of MABNet can be further improved upon better optimization for the values of $\lambda$.

\subsection{Different combinations of features of MABNet}
There exist many possible combinations of settings for our model depending on the different features: (a) The kind of crops (global, local, or both) we feed into the master and the assistant block, (b) the Presence or absence of weight transfer mechanism, (c) Direction of weight transfer if it is present, (d) Presence or absence of self-supervised loss (i.e., KL divergence loss) between the master and the assistant block, (e) Choice of reference in the KL divergence loss.

 We, in our setup, feed global and local crops to the master block and only global crops to the assistant block. We hypothesize that using KL divergence loss by keeping the master block embedding distribution as the reference helps the assistant block learn more detailed information. This further helps the master block become more generalized because of the weight transfer mechanism from the assistant block to the master block. 

In order to justify our hypothesis, we perform a series of experiments exploring all the different types of combinations discussed above. Considering that the master block gives the final output, there must be some learning of the master block from the assistant block. Thus, only those combinations are considered where there is at least one kind of learning support from the assistant to the master: either the parameters of the master block are updated based on the assistant block or the direction of the KL loss is towards the master block, i.e., the assistant embedding is considered as the reference distribution. The results of all the combinations are shown in Table \ref{table:AblationStduy}. The inference is made using only the master block on the Oxford5k dataset with $\lambda = 0.5$. The MABNet, denoted in cyan color, performs better than all the other setups by a significant margin justifying our hypothesis.

\subsubsection{Types of crops}
The global and local crops passed to the master block, and the global crops passed to the assistant block give the best result compared to all the blocks except in some settings where only global crops are passed to the master block, and all the crops are passed to the assistant block. Even when comparing this situation, MABNet performs the best, thus justifying the choice of crops for the master and the assistant block.  

\subsubsection{Weight Transfer and KL Divergence Loss}
Here we analyze the effect of different combinations of presence, absence, and directions of weight transfer and KL divergence loss on the model performance. 

\textbf{Reversed KL divergence direction and absence of Weight Transfer}
We removed weight transfer and reversed the direction of KL Divergence loss to check the importance of the weight transfer mechanism. Even though KL Divergence loss does help the assistant to transfer its knowledge to the master block in the absence of weight transfer, it still leads to the worst performance in almost all cases. This indicates that weight transfer is one of the most important components of our network.

\textbf{Reversing the flow of information} In the original settings of MABNet, we updated the master block using the weighted average of master and assistant weights and used a KL divergence loss between embedding of master and assistant block keeping master block embedding as reference. In this ablation, we study the effect on the model's performance by reversing this flow of information, i.e., updating the assistant block as the average of weights and keeping assistant block embeddings as the reference for KL divergence loss. We see a drop in score to 80.5\%. We hypothesize that the flow of information in MABNet is more effective because we add global-only spatial information of assistance to local-global contextual embeddings of the master block using weight transfer, making it more general.  Moreover, since the master achieves general embeddings, it teaches the assistant block through KL divergence by keeping its embeddings as a reference.

\input{Tables/CombinationOfFeatures}

\section{Conclusion}
We presented MABNet, a new approach to employ both the supervised and self-supervised learning paradigms. Further, as compared to the previously reported teacher-student model, we introduced mechanisms to perform collaborative learning in a guided manner in our buddy model. Our extensive ablation studies clearly show the importance of having the buddy model, guidance of self-supervised learning by the master, and knowledge propagation from assistant to master. 

We show significant improvement in performance over previous approaches and surpass the state-of-the-art mAP scores for both Oxford and Paris datasets as well as their revisited versions. We also show that post-processing improves performance further. Moreover, we set a new SOTA on the Google Landmark dataset, performing at 37\% without post-processing and $\sim$38\% with post-processing using offline diffusion.

In essence, the work presents a powerful learning paradigm that builds on a vision transformer. We expect that the proposed novel paradigms will inspire further developments in vision transformer-based architectures.

We showed how two ViTs with multi-view training help to efficiently leverage local information extracted with a CNN backbone and stressed the superiority of sub-center ArcFace when confronted with long-tailed class distributions and intra-class variability. We confirmed the applicability of the re-ranking and diffusions post-processing techniques to get improvements. The ablation study confirms that the direction of the weight updating and entropy loss between embeddings is crucial to the architecture. Moreover, we got high metrics with concatenated embeddings. The code is available at \textbf{https://github.com/Rohit102497/MABNet}.
\\ \\
{\textbf{Acknowledgement}}
We acknowledge the following funding: Researcher Project for Scientific Renewal grant no. 325741 (Dilip K. Prasad), UiT's thematic project VirtualStain with Cristin ID 2061348 (Alexander Horsch, Dilip K. Prasad) and Horizon 2020 FET open grant OrganVision (id 964800).

\bibliographystyle{IEEEbib}
\bibliography{main}

\end{document}

%% file: Figures/Architecture.tex
\begin{figure*}[t!]
\begin{center}
\includegraphics[width=\textwidth]{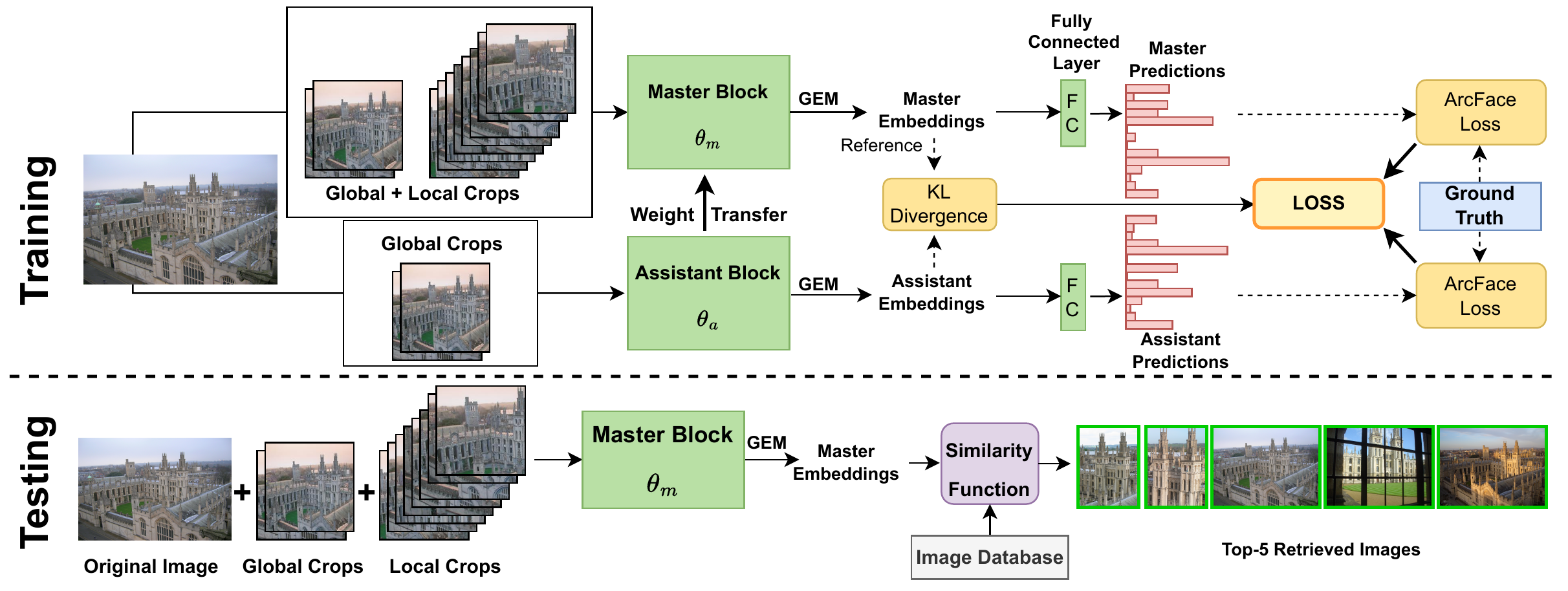}
\end{center}
\vspace{-5mm}
\caption{\textit{Training architecture} (\textit{Upper part}): The architecture of the master assistant buddy network (MABNet). Global crops are fed to assistant block while both local and global crops are used for master block to generate assistant and master embedding, respectively, from Generalized Mean Pooling (GEM) layer. KL divergence loss is calculated over these embeddings. The embedding are then passed through a fully connected layer to get predictions, over which ArcFace loss is applied. After each epoch, the weights of master block are updated by weight transfer.
\textit{Testing architecture} (\textit{Lower part}): The predictions at test time are only performed via the master block. The original input image and its global and local crops are passed to the trained master block to get the embedding from the GEM layer. The concatenation of these embeddings are then compared with the embedding of the images from database with a similarity function (any post-processing method) to retrieve $k$ best images.}
\label{fig:architecture}
\end{figure*}

%% file: Algorithm/Algo1.tex
\begin{algorithm}[t]
\label{algo:Algorithm 1}
\DontPrintSemicolon
\caption{MABNet training algorithm}\label{alg:one}

\SetKwData{Left}{left}\SetKwData{This}{this}\SetKwData{Up}{up}
\SetKwInOut{Input}{Input}\SetKwInOut{Output}{Output}
\SetKwProg{Init}{init}{}{}

\Input{Pretrained Master block: $V_{\theta_m}$, Pretrained Assistant block: $V_{\theta_a}$, Input image: $x$, Ground truth label: $y$, Batch size: $N$, Weight transfer parameter: $\lambda$}
{\textbf{Initialize} : KL Divergence Loss: $L_{KL} = 0$, Assistant ArcFace Loss: $L_{AF}^A = 0$, Master ArcFace Loss: $L_{AF}^M = 0$, Total Loss: $Loss = 0$}\\
\For{$n = 1$ \KwTo $N$}{
    {$C_G$} =  MultiCrop$_G$($x_n$) \tcp*{Global Crops}
    {$C_L$}= {\rm MultiCrop$_L$}($x_n$) \tcp*{Local Crops}
    {$L_{KL} = 0$, $L_{AF}^M = 0$, $L_{AF}^A = 0$}\\
    \For{$i = 1$ \KwTo $|C_G|$}{
        {$E_{a}^{i}$ = $V_{\theta_a}(x_a^i)$; $x_a^i \in \{C_G\}$} \tcp*{Assistant Embeddings}
        {$y^{i}_{a}$} = {\rm $\text{FCLayer}_a$}({$E_{a}^{i}$})\\
        $L_{AF}^{A}$ += {\rm ArcFaceLoss}($y^{i}_{a}$, $y_n$)
    }
    \For{$j = 1$ \KwTo $|C_G \cup C_L|$}{
        {$E_{m}^{j}$ = $V_{\theta_m}(x_m^j)$;  $x_m^j \in \{C_G \cup C_L\}$ } \tcp*{Master Embeddings}
        {$y^{j}_{m}$} = {\rm $\text{FCLayer}_m$}({$E_{m}^{j}$})\\
        $L_{AF}^{M}$ += \rm ArcFaceLoss($y^{j}_{m}$, $y_n$)
    }
    \For{$i \in \{1,...,|C_G| \}$ and $j \in \{1,...,|C_G \cup C_L| \}$}{
        $L_{KL}$ += \rm KLLoss($E_{a}^{i}$, $E_{m}^{j}$)
    }
    $Loss + = L_{KL} + L_{AF}^{M} + L_{AF}^{A}$
}
{$\frac{\partial Loss}{\partial \theta_m}, \frac{\partial Loss}{\partial \theta_a}$}
\tcp*{Backward pass updating parameters $\theta_m$ and $\theta_a$}
{$\theta_m$ = $\lambda*\theta_m$ + $(1-\lambda)\theta_a$}  
\tcp*{Weight Transfer}
\end{algorithm}

%% file: Algorithm/Algo2.tex
\begin{algorithm}[t]
\label{algo:Algorithm2}
\DontPrintSemicolon
\caption{MABNet testing algorithm}\label{alg:two}

\SetKwData{Left}{left}\SetKwData{This}{this}\SetKwData{Up}{up}
\SetKwInOut{Input}{Input}\SetKwInOut{Output}{Output}

\Input{Fully trained Master block: $V_{\theta_m}$, Images in database: $X$, Query image: $q$, Number of images to be retrieved: $k$}
\Output{Top $k$ images similar to the query image}

\For{$x_i\in X$}{
$E_i$ = $V_{\theta_m}$($x_i$) \tcp*{Creating Database Embeddings}}
E = [$E_1$, $E_2$, .... $E_{|X|}$]\\
$E^q$ = $V_{\theta_m}$($q$) \tcp*{Query Embedding}
Output = SimilarityFunc($E^q$, $E$, $k$)\\
where SimilarityFunc is postprocessing methods discussed in section \ref{sec:PostProcessingMethods}
\end{algorithm}

%% file: Figures/FigRetreivalExample.tex
\begin{figure}[t]
\begin{center}
\includegraphics[width=\linewidth]{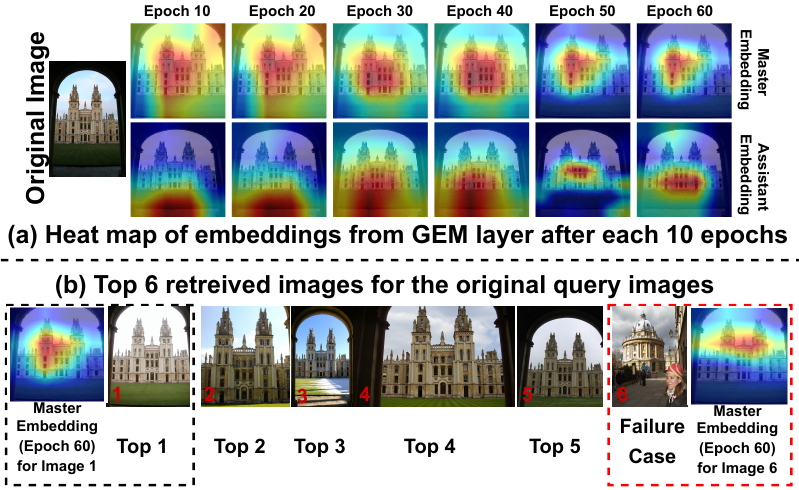}
\end{center}
\vspace{-5mm}
\caption{ (a) The superimposed heat map of the the original image after every 10 epochs from both the master and the assistant embedding is generated using Grad-CAM till 60 epochs.
(b) Top 5 retrieved image for the original query image. We also show the top 6 retrieved image which is a failure case. The left and right heat map is from the master embedding for the top 1 retrieved image and the failure case, respectively. This result is on Ox105k dataset without post-processing.}
\label{fig:imageretrievalexample}
\vspace{-3mm}
\end{figure}

%% file: Tables/WithoutPostProcessing.tex
\setlength{\tabcolsep}{4.5pt}
\begin{table}[t]
\begin{center}
\caption{mAP scores of various methods without post-processing on Oxford, Paris and their revisited datasets. MABNet-R50 represent MABNet with ResNet-50 backbone.}
\label{table:resultsw/opostprocessing}
\begin{tabular}{lcccc}
\hline\noalign{\smallskip}
Model & Ox5k & Ox105k & P6k & P106k\\
\noalign{\smallskip}
\hline
R-MAC \cite{DBLP:journals/corr/ToliasSJ15} & 66.9 & 61.6 & 83.0 & 75.7\\
siaMAC \cite{radenovic2016cnn} & 80.0 & 75.1 & 82.9 & 75.3\\
DIR \cite{gordo2016deep} & 83.1 & 78.6 & 87.1 & 79.7\\
DELF \cite{noh2017large} & 83.8 & 82.6 & 85.0 & 81.7\\
Deep Conv\cite{hoang2017selective} & 83.8 & 80.6 & 88.3 & 83.1\\
\rowcolor{LightCyan} MABNet & \textbf{91.5} & \textbf{88.7} & \textbf{94.2} & \textbf{90.4}\\
\hline
\noalign{\smallskip}
& ROx(M) & ROx(H) & RPar(M) & RPar(H)\\
\noalign{\smallskip}
\hline
DINO \cite{caron2021emerging} & 51.5 & 24.3 & 75.3 & 51.6\\
IRT(R) \cite{elnouby2021training} & 55.1 & 28.3 & 72.7 & 49.6 \\ 
Listwise\cite{revaud2019learning} & 67.5 & 42.8 & 80.1 & 60.5 \\ 
SOLAR \cite{ng2020solar} & 69.9 & 47.9 & 81.6 & 64.5\\
\rowcolor{LightCyan} MABNet & \textbf{85.0} & \textbf{61.2} & \textbf{84.6} & \textbf{72.9}\\
MABNet-R50 & 80.3 & 59.2 & 82.0 & 67.4\\
\hline
\end{tabular}
\end{center}
\vspace{-4mm}
\end{table}
\setlength{\tabcolsep}{4.5pt}

%% file: Tables/WithPostProcessing.tex
\setlength{\tabcolsep}{1.4pt}
\begin{table}[t]
\begin{center}
\caption{mAP scores of various methods with different post-processing techniques on Oxford, Paris and their revisited datasets. Legend: PP - PostProcessing, Q - AQE, AQ - AML + AQE,  DQ - DIR + AQE, WQR - HeW + AQE + HeR, O - Offline Diffusion, CV - CVNet.}
\label{table:resultswpostprocessing}
\begin{tabular}{lcccc}
\hline\noalign{\smallskip}
Model + PP & Ox5k & Ox105k & P6k & P106k\\
\noalign{\smallskip}
\hline
\rowcolor{LightCyan} MABNet  & 91.5 & 88.7 & 94.2 & 90.4 \\
\hline
siaMAC + Q \cite{radenovic2016cnn} & 85.4 & 82.3 & 87.0 & 79.6\\
DIR + Q \cite{gordo2016deep} & 89.0 & 87.8 & 93.8 & 90.5 \\
\rowcolor{LightCyan} MABNet + Q & 92.3 & 89.6 & 95.2 & 91.1 \\
\hline
R-MAC + AQ \cite{DBLP:journals/corr/ToliasSJ15} & 77.3 & 73.2 & 86.5 & 79.8\\
DELF + DQ \cite{noh2017large} & 90.0 & 88.5 & 95.7 & 92.8 \\
siaMAC+WQR\cite{pang2018deep} & 92.0 & 90.3 & 94.3 & 90.2 \\
\rowcolor{LightCyan} MABNet + WQR & 95.7 & 92.9& 96.5& 95.6 \\
\hline
R-MAC + O \cite{yang2019efficient} & 96.2 & 95.2 & 97.8 & 96.2 \\
\rowcolor{LightCyan} MABNet + O & \textbf{97.2} & \textbf{96.3} & \textbf{98.1} & \textbf{96.8} \\
\hline
\hline
\noalign{\smallskip}
& ROx(M) & ROx(H) & RPar(M) & RPar(H)\\
\noalign{\smallskip}
\hline
\hline
DELG \cite{cao2020unifying} & 81.2 & 64.0 & 87.2 & 72.8 \\
DOLG \cite{yang2021dolg} & 81.5 & 61.1 & 91.0 & 80.3\\
Swin-T-DALG \cite{song2022dalg} & 78.7 & 54.7 & 88.2 & 76.3\\
Swin-S-DALG \cite{song2022dalg} & 79.9 & 57.5 & 90.4 & 79.0\\
GeM (Baseline) \cite{hu2022expressive} & 83.0 & 65.5 & 90.2 & 80.7\\
GeM-Local Match \cite{hu2022expressive} & 85.9 & 71.2 & 92.0 & \textbf{83.7}\\
CVNet-Rerank \cite{Lee_2022_CVPR} & 87.2 & 75.9 & 91.2 & 81.1\\
\hline
MABNet + Q & 87.1 & 63.8 & 86.4 & 73.5 \\
MABNet + WQR & 88.4 & 65.4 & 88.3 & 75.8 \\
MABNet + O & \textbf{89.3} & 66.2 & 88.9 & 78.2 \\
\rowcolor{LightCyan} MABNet + CV & 87.1 & \textbf{76.5} & \textbf{92.3} & 82.7\\
\hline
\end{tabular}
\end{center}
\vspace{-4mm}
\end{table}
\setlength{\tabcolsep}{1.4pt}

%% file: Tables/GLDv2Dataset.tex
\setlength{\tabcolsep}{4pt}
\begin{table}[t]
\begin{center}
\caption{mAP score of MABNet with and without post processing vs other models on test images of GLDv2 dataset.}
\label{table:gldv2benchmarking}
\begin{tabular}{llc}
\hline\noalign{\smallskip}
Model & Post Processing & GLDv2\\
\noalign{\smallskip}
\hline
\noalign{\smallskip}
ResNet101+ArcFace \cite{weyand2020google} &  & 28.5 \\
Zhang et al. \cite{yuqi20212nd} &  & 32.4 \\
{DELG} \cite{cao2020unifying} &  & 26.8 \\
\rowcolor{LightCyan} MABNet &  & \textbf{37.0} \\
\hline
MABNet & AQE & 37.2 \\
MABNet & HeW+AQE+HeR & 37.8 \\
MABNet & Offline Diffusion & 38.1 \\

\hline
\end{tabular}
\end{center}
\end{table}
\setlength{\tabcolsep}{1.4pt}

%% file: Tables/Ensemble.tex
\setlength{\tabcolsep}{1.4pt}
\begin{table}[t]
\begin{center}
\caption{mAP scores of MABNet(C) without post-processing on Oxford, Paris and their revisited datasets.}
\label{table:resultensemble}
\begin{tabular}{lcccc}
\hline{\smallskip}
Model & Ox5k & Ox105k & P6k & P106k\\
\noalign{\smallskip}
\hline
MABNet(C) & 92.5 & 89.6 & 95.3 & 92.1 \\
MABNet(C) + Q & 93.6 & 92.4 & 95.9 & 94.1\\
MABNet(C) + WQR & 97.1 & 93.8 & 98.1 & 96.4\\
MABNet(C) + O & 98.3 & 97.8 & 98.3 & 97.5\\
\hline
\noalign{\smallskip}
& ROx(M) & ROx(H) & RPar(M) & RPar(H)\\
\noalign{\smallskip}
\hline
MABNet(C) & 85.8 & 62.1 & 85.7 & 73.8 \\
MABNet(C) + Q & 88.1 & 62.8 & 86.2 & 74.7\\
MABNet(C) + WQR & 89.7 & 65.8 & 87.6 & 76.9\\
MABNet(C) + O & 90.2 & 67.5 & 89.8 & 79.4\\
\hline
\end{tabular}
\end{center}
\vspace{-4mm}
\end{table}
\setlength{\tabcolsep}{1.4pt}

%% file: Tables/LambdaAblation.tex
\setlength{\tabcolsep}{4pt}
\begin{table}[t]
\begin{center}
\caption{mAP scores for different values of $\lambda$. All the results reported here are without any pre-processing techniques.}
\label{table:LambdaAblation}
\begin{tabular}{lccccc}
\hline\noalign{\smallskip}
$\lambda$ & Ox5k & ROx(M) & ROx(H) & RPar(M) & RPar(H)\\
\noalign{\smallskip}
\hline
\noalign{\smallskip}
0.30 & 75.9 &79.9 &57.6 &80.4 &68.5\\
0.35 & 81.8 &81.5 &57.3 &81.6 &71.2\\
0.40 & 87.4 &84.3 &59.4 &84.1 &70.8\\
0.45 & 90.2 &85.8 &60.3 &83.2 &72.5\\
0.50 & 91.5 & 85 & 61.2 & 84.6 & 72.9\\
0.55 & 92.7 & \textbf{87.3} & 64.5 & 85.4 & \textbf{75.2}\\
0.60 & \textbf{93.2} & 86.4 & \textbf{67.8} & 86.7 & 74.6\\
0.65 & 92.9 & 85.8 & 66.2 & \textbf{87.3} & 73.7\\
0.70 & 92.0 & 83.4 & 60.6 & 85.2 & 72.5\\
0.75 & 90.6 & 82.2 & 59.2 & 84.5 & 71.3\\
0.80 & 88.1 & 76.5 & 58.4 & 83.2 & 70.8\\
0.85 & 79.6 & 75.3 & 57.6 & 81.4 & 68.5\\
0.90 & 78.5 & 72.1 & 55.9 & 80.9 & 66.7\\
0.95 & 78.2 & 70.2 & 53.2 & 79.5 & 65.2\\
\hline
\end{tabular}
\end{center}
\end{table}
\setlength{\tabcolsep}{1.4pt}

%% file: Tables/CombinationOfFeatures.tex
\setlength{\tabcolsep}{6pt}
\begin{table}[!t]
\vspace{-2mm}
\centering
\caption{Results for different permutations and combinations of the features of MABNet are shown here. Here, we use acronyms to denote some variables. KL denotes the  Kullback-Leibler loss, WT denotes the weight transfer, mAP denotes the mean average precision of the model, Master crops denote the crops fed to the master block, and Assistant crops denote the crops fed to the assistant block.
The meaning of the notations used in the table is: G represents global crops, L represents local crops, X means this feature is not present in the model, $\uparrow$ arrow means the direction of the feature is upward, i.e., if KL is up then the assistant embedding is taken as the reference, and $\downarrow$ arrow means the direction is downwards, i.e., if the WT is down, it means the parameters of the assistant block gets updated as a weighted average. The MABNet (denoted in Cyan) performs best compared to other settings. The result shown here is based on the Oxford5k dataset without any post-processing techniques.}
\label{table:AblationStduy}
\begin{tabular}{ccccc}
\hline
\noalign{\smallskip}
Master crops & Assistant crops & KL & WT & mAP\\
\noalign{\smallskip}
\hline
\noalign{\smallskip}

G+L & G & X & $\uparrow$ & 83.3\\
\rowcolor{LightCyan}    &   & $\downarrow$ & $\uparrow$ & \textbf{91.5}\\
    &   & $\uparrow$ & X & 64.1\\
    &   & $\uparrow$ & $\downarrow$ & 80.5 \\
    &   & $\uparrow$ & $\uparrow$ & 85.7\\
\hline

G+L & L & X & $\uparrow$ & 61.9\\
    &   & $\downarrow$ & $\uparrow$ & 64.2\\
    &   & $\uparrow$ & X & 31.7 \\
    &   & $\uparrow$ &$\downarrow$ & 61.6 \\
    &   & $\uparrow$ & $\uparrow$ & 57.8 \\
\hline

G & G+L & X & $\uparrow$ & 69.2\\
    &   & $\downarrow$ & $\uparrow$  & 73.3\\
    &   & $\uparrow$ & X & 69.7\\
    &   & $\uparrow$ & $\downarrow$ & 84.2\\
    &   & $\uparrow$ & $\uparrow$ & 76.1\\
\hline

G & L & X & $\uparrow$ & 52.5\\
    &   & $\downarrow$ & $\uparrow$ & 49.3\\
    &   & $\uparrow$ & X & 31.9\\
    &   & $\uparrow$ & $\downarrow$ & 31.4\\
    &   & $\uparrow$ & $\uparrow$ & 42.6\\
\hline

L & G+L & X & $\uparrow$ & 55.4\\
    &   & $\downarrow$ & $\uparrow$ & 52.7\\
    &   & $\uparrow$ & X & 30.5\\
    &   & $\uparrow$ & $\downarrow$ & 48.7\\
    &   & $\uparrow$ & $\uparrow$ & 34.6\\
\hline

L & G & X & $\uparrow$ & 39.9\\
    &   & $\downarrow$ & $\uparrow$ & 29.6\\
    &   & $\uparrow$ & X & 25.2\\
    &   & $\uparrow$ & $\downarrow$ & 23.4\\
    &   & $\uparrow$ & $\uparrow$ & 30.1\\
\hline
\end{tabular}
\end{table}